%% file: iccv.tex
\documentclass[10pt,twocolumn,letterpaper]{article}

\usepackage{iccv}
\usepackage{times}
\usepackage{epsfig}
\usepackage{graphicx}
\usepackage{amsmath}
\usepackage{amssymb}
\usepackage{booktabs}
\usepackage{tabularx}
    \newcolumntype{L}{>{\raggedright\arraybackslash}X}
\usepackage[font=small]{caption}
\usepackage{subcaption}
\usepackage{stmaryrd}
\usepackage{stfloats}    
\usepackage{pifont}
\newcommand{\cmark}{\ding{51}}%
\newcommand{\xmark}{\ding{55}}%
\usepackage{authblk}

\usepackage[pagebackref=true,breaklinks=true,letterpaper=true,colorlinks,bookmarks=false]{hyperref}

\iccvfinalcopy 

\def\iccvPaperID{8737} 

\ificcvfinal\fi

\newcommand{\R}{\mathbb{R}}
\DeclareMathOperator*{\argmin}{arg\,min}
\renewcommand{\vec}[1]{\mathbf{#1}}

\begin{document}

\makeatletter
\renewcommand\AB@affilsepx{, \protect\Affilfont}
\makeatother

\title{UVStyle-Net: Unsupervised Few-shot Learning of 3D Style Similarity Measure for B-Reps}

\author[1,3]{Peter Meltzer}
\author[1,3]{Hooman Shayani}
\author[1]{Amir Khasahmadi}
\author[2]{Pradeep Kumar Jayaraman}
\author[1]{Aditya Sanghi}
\author[1]{Joseph G. Lambourne}
\affil[1]{Autodesk AI Lab}
\affil[2]{Autodesk Research}
\affil[3]{UCL}

\maketitle
\ificcvfinal\thispagestyle{empty}\fi

\begin{figure*}[b]
\begin{center}
   \includegraphics[width=.9\linewidth]{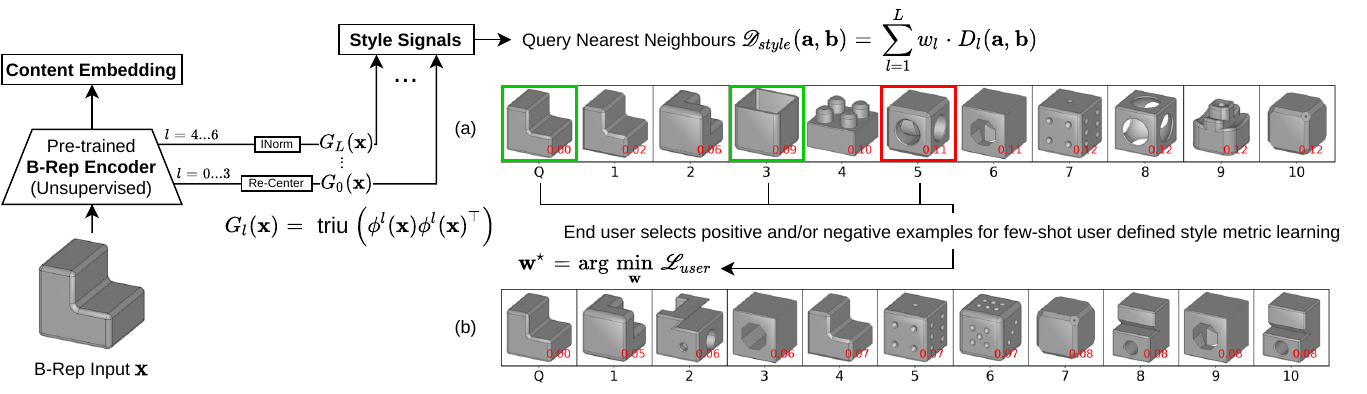}
\end{center}
   \caption{Overview of UV-StyleNet: Grams of activations are normalized and extracted for each layer. The weights applied to each layer define the meaning of style. (a) Top-10 query results using uniform layer weights $\vec{w}$ (b) Top-10 query results using $\vec{w}^\star$ based on the user-selected examples (positive in green, negative in red). In this example, $\vec{w}^\star \approx [0, 0, 0, 1, 0, 0, 0]^\top$. Zoom to see fillets/stylistic details.}
\label{fig:overview}
\end{figure*}

\begin{abstract}
   \input{abstract}
\end{abstract}

\input{body}

{\small
\bibliographystyle{ieee_fullname}
\bibliography{iccv.bib}
}

\input{appendix}

\end{document}

%% file: abstract.tex
Boundary Representations (B-Reps) are the industry standard in 3D Computer Aided Design/Manufacturing (CAD/CAM) and industrial design due to their fidelity in representing stylistic details. However, they have been ignored in the 3D style research. Existing 3D style metrics typically operate on meshes or point clouds, and fail to account for end-user subjectivity by adopting fixed definitions of style, either through crowd-sourcing for style labels or hand-crafted features. We propose UVStyle-Net, a style similarity measure for B-Reps that leverages the style signals in the second order statistics of the activations in a  pre-trained (unsupervised) 3D encoder, and learns their relative importance to a subjective end-user through few-shot learning. Our approach differs from all existing data-driven 3D style methods since it may be used in completely unsupervised settings, which is desirable given the lack of publicly available labeled B-Rep datasets. More importantly, the few-shot learning accounts for the inherent subjectivity associated with style.
We show quantitatively that our proposed method with B-Reps is able to capture stronger style signals than alternative methods on meshes and point clouds despite its significantly greater computational efficiency. We also show it is able to generate meaningful style gradients with respect to the input shape, and that few-shot learning with as few as two positive examples selected by an end-user is sufficient to significantly improve the style measure. Finally, we demonstrate its efficacy on a large unlabeled public dataset of CAD models. Source code and data are available at \href{https://github.com/AutodeskAILab/UVStyle-Net}{github.com/AutodeskAILab/UVStyle-Net}.

%% file: body.tex
\section{Introduction}

B-Reps are the de facto standard for industrial design, and the representation most widely used in the consumer product and automotive industries where style is of great importance. B-Reps offer unparalleled editability in a compact, memory efficient representation, they are not discrete/sampled (as per mesh/point cloud) offering precise boundaries with continuous smooth surfaces/edge curves. See \autoref{app:breps} for a brief introduction to B-Reps. A B-Rep style similarity measure has many use cases, \ie finding architectural parts that are in-keeping with the style of a building, or selecting parts for a car that fit with the manufacturer's existing range. Moreover, the gradient of a style similarity measure can be used to generate helpful visualizations or modify the input 3D shape à la Gatys \etal \cite{Gatys2016}.

Geometric style is inherently subjective and may have a different meaning in different object class domains, \ie the boundary between style and content is unclear. For example, in the context of chair designs, number of legs could be considered either style or content depending on the particular use case. Thus, an effective geometric style measure must cater for these different interpretations of the end user.

While existing methods use hand-crafted features \cite{Lun2015, Liu2015} or crowd-sourcing \cite{Lim2016, Pan2017, Polania2020, Pan2019} to pre-define and measure geometric style, we propose a user-defined few-shot style metric learning method that leverages the range of style signals available in the activations of a pre-trained 3D object encoder through second order statistics (Gram matrices). The relative importance of each layer's Gram matrix is then learnt through selection of just a few examples of what style means to an end user (see \autoref{fig:overview}).

\begin{figure}[t]
\begin{center}
   \includegraphics[width=.7\linewidth]{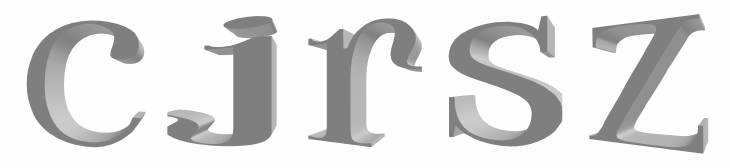}
\end{center}
   \caption{Lower case examples from font `Viaoda Libre'. While `j' and `r' share some stylistic features, they are not obviously similar to `c', `s' or `z', \ie font classes provide a ground truth for style compatibility (as perceived by their designers) yet only a \emph{weak} label for style itself.}
\label{fig:viaoda_libre_egs}
\end{figure}

Despite the abundant use of B-Reps in industrial settings, there is a fundamental lack of publicly available B-Rep data for training machine learning models --- in particular, there are no existing B-Rep datasets that include a reliable ground truth for style. To overcome this challenge, we provide an adaptation to SolidLetters \cite{Jayaraman2020}, which improves the style consistency within font classes for the evaluation test set. The font classes, however, still provide only a weak label for style (see \autoref{fig:viaoda_libre_egs}), and as such we propose an unsupervised method and use the font labels purely for quantitative evaluation to justify design choices of our method. For comparison against existing SOTA on real-world data we also provide evaluation with the unlabeled ABC dataset \cite{KochBerlinskoch} of CAD models and a manually labeled subset of it.

The main contributions of this work are as follows:
\begin{itemize}
    \item We demonstrate that the second order statistics (Gram matrix) approach used in 2D image style literature can be generalised to (B-Rep) 3D shapes
    \item We introduce a general few-shot learning method for capturing a subjective end-user's definition of 3D style and demonstrate its effectiveness on B-Reps
    \item We show quantitative efficiency and performance advantages of using UVStyle-Net architecture with B-Reps over similar approaches on meshes and point clouds using a new synthetic public dataset (SolidLetters) and a small subset of ABC labeled for style
    \item We verify our method on the ABC dataset with no style or content labels for pre-training, and demonstrate the effectiveness of our few-shot learning process to capture subjective user-defined style similarity measures
\end{itemize}

In summary, we introduce a geometric style similarity measure for 3D solids that may be used in completely unlabeled settings for arbitrary object classes, with user subjectivity handled by few-shot learning given only a few examples. While our method is adaptable for all 3D input types, we demonstrate the benefits of our approach with B-Reps both quantitatively and qualitatively.
\section{Related Work}

\textbf{Geometric Feature Learning.} Geometric feature learning has seen many successes for both Euclidean representations, \ie multi-view \cite{Su}, projections \cite{Cao}, volumetric \cite{ZhirongWu2015}, and non-Euclidean representations, \ie point clouds \cite{Wang2019b,Qi2017,Guo2019} and mesh \cite{Hanocka2019,DeHaan2020}. For a detailed review of geometric feature learning we refer the reader to \cite{Bronstein2017,Griffiths2019,Ahmed2019}. Despite the prevalence of B-Reps in industrial and creative design applications, however, geometric feature learning for parametric representations remains largely unexplored.

In addition to their wide use, there are many advantages to working with B-Reps as 3D geometric representations. Not only do B-Reps typically require less memory than point clouds or meshes (depending on the sampling resolution/detail of the model), but they also provide richer information about a solid, including the precise boundaries of every surface and the topology of these surfaces.

The benefits of B-Reps over discrete representations are demonstrated in  Jayaraman \etal \cite{Jayaraman2020}, where each face is sampled uniformly in its parameter domain to form a regular grid then passed through a 2D CNN. The CNN face representations are then fed to a Graph Neural Network (GNN) which uses the face adjacency matrix of the original B-Rep.

\textbf{Geometric Style Similarity.} Existing geometric style similarity learning methods are typically trained in a supervised setting, requiring a set of hand-labeled triplets in which one pair is believed to be closer in style than the other \cite{Liu2015,Lun2015,Lim2016,Pan2017,Pan2019,Polania2020}. To account for style subjectivity, examples are labeled through crowd-sourcing methods and thus result in a generally accepted definition for style. For example, Liu \etal \cite{Liu2015} use hand-crafted features (\ie curvature histograms) with a supervised triplet loss to learn furniture compatibility, while Lun \etal \cite{Lun2015} apply a similar method by first segmenting input models into sub-parts to compute geometric features for independently.

For geometric style feature learning, Lim \etal \cite{Lim2016} and Pan \etal \cite{Pan2017} project 3D meshes into multiple 2D views which are fed into a triplet image CNN. Polania \etal \cite{Polania2020} adopt a similar approach, where the learned style representations are then passed to a GNN for compatibility prediction. Rendering 3D solids into 2D (even with multiple views) is problematic since stylistic features can be lost or occluded and selecting the best views without making assumptions on the orientations of the data is non-trivial. Pan \etal \cite{Pan2019} overcome this using curvature-guided sampling directly from the solids to generate element-level style features which are then aggregated to global style representations using a triplet network.

The reliance of these methods on crowd-sourced, hand-labeled style triplets creates two problems: Firstly, there is limited labeled data available in the 3D style domain, and no labeled B-Rep data. Secondly, and more importantly, the definition of style (an inherently subjective concept) is pre-defined according to a consensus, hence may not be compatible with an end-user's particular taste or application.

\textbf{Style Transfer.} Contrary to the geometric style learning methods above, the style transfer literature has largely adopted the use of first and second order activation statistics from deep pre-trained image classifiers in order to represent and quantify style. Gatys \etal \cite{Gatys2016} showed that feature co-occurrence in the different layers of a CNN effectively captures elements of style at different scales of abstraction. In the finest layers where features are most local, the style representation given by the Gram matrix captures colour and texture information, yet deeper into the network, the Gram matrices capture higher level structure and patterns eventually crossing into semantic content. Following from this, Huang \etal \cite{Huang2017} and Babaeizadeh \etal \cite{Babaeizadeh2019} demonstrate that first order activation statistics (channel-wise mean and variance) can also capture elements of style through the use of Adaptive Instance Normalization (AdaIN). Karras \etal \cite{Karras2019} illustrate the relationship between layer depth and the style/content trade-off by swapping the inputs to a generator at varied depth. Swapping at lower layers renders image interpolations of low level texture/colour information, and swapping at deeper layers interpolates semantic content.

Many further works utilise and extend the use of first order  statistics of network activations to improve style transfer results, e.g. GAN based methods \cite{Lehtinen,Zhanga,Jiang2020}; however, these methods rely on a generator to align the activations to these statistics while generating an output image, mainly focusing on the quality of the output images rather than the interpretability of the statistics in defining an explicit style distance metric for arbitrary inputs. To explicitly disentangle style and content for arbitrary inputs \cite{Park2020} proposes an auto-encoder that adopts the technique of swapping inputs at various layers and a GAN based encoder and discriminator that is able to effectively separate structure and texture.

Azadi \etal \cite{Azadi} propose a few-shot learning approach for font style transfer in which stacked conditional GANs are used to generate unseen characters in a target style from a small number of observed examples. This method is, however, specific to font generation and relies on supervised pre-training using the style labels.

\textbf{3D Style Transfer.} Recently, Liu \etal \cite{DerekLiu2020} showed that style could be learned from one mesh model and transferred to another using a neural subdivision surface scheme. Cao \etal \cite{Cao2020} generalised the second order statistics approach of \cite{Gatys2016} to 3D point clouds, adopting the use of a Pointnet \cite{Qi} encoder pre-trained for classification on ShapeNet \cite{Chang2015}. Following the trend of 2D style transfer Segu \etal \cite{Segu} extend this work using GAN methods to produce a generative model with better disentanglement of content and style. There are no existing style transfer/unsupervised approaches to style metric learning for B-Reps.
\section{UVStyle-Net}

Inspired by image-based style-transfer, our method uses second order statistics of the activations from a pre-trained B-Rep encoder to form a flexible style representation.

For the encoder we use UV-Net \cite{Jayaraman2020}, which processes each face of a solid with 3 layers of 2D convolutions, and propagates the projected pooled features of each face in a face-adjacency graph using 2 GIN \cite{Xu} layers. Each face is represented by a $10 \times 10$ grid (image) of 7 dimensions containing the absolute 3D position (xyz) of each UV sample, the normal for each sample, and a mask indicating whether each sample lies within or outside of the trimmed face. We use UV-Net due to its SOTA performance on B-Rep classification and its parallels to conventional 2D CNNs.

For B-Rep model $\vec{x}$, we extract the normalised, flattened upper triangle of the Gram matrix (modelling the feature correlations \cite{Gatys2016}) for each layer $l$:
\begin{equation}
\label{eq:gram}
G_l\left(\vec{x}\right) =
    \text{triu}\left(
        \phi^{l}(\vec{x}) \phi^l(\vec{x})^\top
    \right)
\end{equation}
where $\phi^l(\vec{x}) \in \R^{d_l \times N_l}$ is the normalised feature map of a pre-trained classifier given input $\vec{x}$ such that $\phi^l_{ij}(\vec{x})$ is the normalised activation of filter $i$ at position $j$ in layer $l$, $d_l$ and $N_l$ are the number of distinct filters and non-masked samples in layer $l$ respectively,
 and $\text{triu}: \R^{d_l \times d_l} \rightarrow \R^{\frac{d_l (d_l + 1)}{2}}$ returns the flattened upper triangle of a matrix.

For the first (features) layer, samples corresponding to the positions that do not lie on the surface of a trimmed face are masked, and the gram matrix is calculated accordingly.
In the GIN layers, we have a single vector per face (\ie node), thus instance normalization \cite{Ulyanov2017} is applied across the solid prior to computing the Grams. For each of the features (non-masked positions and normals) and activations of each convolution layer's filters, we leverage the grouping of samples into faces which is unique to B-Reps (compared to meshes and point clouds), whereby we re-center (subtract the mean of) the UV samples by face. This can be interpreted as per-face instance normalization without division by the standard deviation.

Face re-centering/instance normalization are applied to the activations after extraction from the encoder, but the raw (un-normalised) activations are passed to the next layer of the encoder, thus imposing no requirements on the encoder architecture in terms of normalization strategies.

Analogous to style-transfer with 2D images \cite{Gatys2016}, for a pair of B-Reps $\vec{a}$ and $\vec{b}$ we define the style distance:
\begin{equation}
\label{eq:style_distance}
\mathcal{D}_{style}\left(\vec{a}, \vec{b}\right) = \sum_{l=1}^L w_l \cdot D_l\left(\vec{a}, \vec{b}\right),
\end{equation}
where
\begin{equation}
D_l\left(\vec{a}, \vec{b}\right) =
    1 - \frac{G_l(\vec{a}) \cdot G_l(\vec{b})}
{\lVert G_l(\vec{a}) \rVert \lVert G_l(\vec{b}) \rVert}
\end{equation}
and $\vec{w}$ is a weights vector that controls how much each layer contributes to the style distance measure. We deviate from Gatys \etal \cite{Gatys2016} in use of the cosine distance (rather than Euclidean) due to simplified normalization and an observed improvement in our initial experiments.

Given a set of user selected examples from a target style (\ie positive samples) $T$, and a set of user selected counter-examples (\ie negative samples) $T'$, we define the user-defined loss:

\begin{equation}
\label{eq:user_loss}
\mathcal{L}_{user} = \sum_{l=1}^L w_l \cdot E_l
\end{equation}
where

\begin{equation}
\label{eq:layer_energy}
E_l = c_1 \cdot \sum_{\substack{\vec{t}_i, \vec{t}_j \in T\\ i \ne j}} D_l(\vec{t}_i, \vec{t}_j)
- c_2 \cdot \sum_{(\vec{t}, \vec{t}') \in T \times T'} D_l(\vec{t}, \vec{t}')
\end{equation}
is a layer-wise energy term, $c_1$ and $c_2$ are normalization constants, and to prevent trivial solutions $\vec{w}$ is constrained such that $\sum_{l=1}^L w_l = 1$ and $\vec{w} \succeq 0$.
Due to these constraints, we note that even with only positive examples $T$ (\ie $T' = \emptyset$), $E_l$ is sufficiently determined, and in such a case the second term may be omitted. However, to reduce the risk of overfitting, a large number of negative examples may be drawn randomly from the remaining dataset. This is of particular benefit in real world settings without access to labeled datasets, where an end user may select only a handful of positive examples that share style as they perceive it.

We find the optimal weights for an end-user
\begin{equation}
\label{eq:convex}
    \vec{w}^\star = \argmin_{\vec{w}} \sum_{l=1}^L w_l \cdot E_l
\end{equation}
subject to the above constraints, and substitute them into Eq. (\ref{eq:style_distance}) to produce the final user style distance metric.

We observe that $E_l$ is constant w.r.t. $\vec{w}$, thus Eq. (\ref{eq:convex}) is simply a linear combination and its intersection with the hyperplane $\sum_{l=1}^L w_l = 1$ results in a twice-differentiable convex optimization which we solve using Sequential Least Squares Quadratic Programming (SLSQP) \cite{Virtanen2020}.

\textbf{Intuition Behind the Grams.} At 0\_feats (inputs), the Gram matrix models the distribution of the position and surface normal of the sampled points with 2nd order statistics. Based on the CNN receptive fields, we understand that the 1st layer Gram is modeling the distribution of local curvatures (\ie flat/saddle/doubly curved), and the next levels capturing those of higher order curvatures (\ie s-shaped), then leading into correlations of patterns of these lower level features, and eventually into content.
\section{Experiments \& Results}

We first test if a method similar to image style approaches can capture 3D style, and quantify the presence of this signal at each layer. We evaluate our method for disentanglement of style from content with a gradient visualization, thus demonstrating a practical use-case in which a designer may utilise the feedback from the model. We then test few-shot learning of our style metric in its ability to capture an end-user's subjective requirements. Finally, we assess the effectiveness of our
completely unsupervised encoder pre-training approach without content labels.

For data, we start with SolidLetters \cite{Jayaraman2020}, which is a collection of extruded letters from a variety of fonts including labels for both content (\ie letter class) and style (\ie font class) (\autoref{tab:solid_mnist}). This is a good choice of data for initial validation of our design decisions, as the 2D nature of the elements of style in 3D shapes simplifies the analysis and debugging while the generation process of these 3D letters mirrors the most typical CAD modelling approach --- drawing a 2D wire body, then extruding to 3D and potentially filleting/bevelling the edges. Following this, we use the real-world ABC dataset \cite{KochBerlinskoch} of CAD models.

In all cases on SolidLetters we pre-train the classifier on the training set to predict the letter, and perform model selection for the best classifier with the validation set. Following the methodology of Cohen \etal \cite{Cohen2017} and Jayaraman \etal \cite{Jayaraman2020}, we perform pre-training using 26 classes (combining upper and lower case examples). SolidLetters includes randomness in the fillet size, and extrusion depth and angle, so for the held-out test set used in all our evaluations, we regenerate the letters to remove sources of randomness (extrusion angle/amount and fillet radius) within font classes, hence strengthening the style labels. For further detail, see \autoref{app:generation}. After pre-training, all experiments are performed using the held-out test set. We note in particular that no examples of the test fonts are included in the training/validation sets, and that font style labels are used purely for evaluation and not during pre-training. For all experiments on ABC, we perform unsupervised pre-training using point cloud reconstruction on the complete dataset.

\begin{table}[htb]
    \centering
    \begin{tabularx}{\linewidth}{Xrrr}
        \toprule
        {}              & Train     & Validation & Test \\
        \midrule
        Examples        & 40,402    & 10,100    & 13,339 \\
        Letter Classes  & 26        & 26        & 26 \\
        Font Classes    & 1,664     & 1650      & 378 \\
        Random Extrude/Fillet & \cmark & \cmark & \xmark \\
        \bottomrule
        \end{tabularx}
    \caption{Details of SolidLetters dataset \cite{Jayaraman2020}. The test set is regenerated without sources of randomness within font classes to strengthen the associated style labels used for evaluation.}
    \label{tab:solid_mnist}
\end{table}

For comparison with other representations and encoders, we use MeshCNN \cite{Hanocka2019}, and Pointnet++ \cite{Qi2017}. We use Pointnet++ over DGCNN \cite{Wang2019b} or Pointnet \cite{Qi} since we are drawing upon 2D style literature. DGCNN aggregates intermediate layer activations according to locality in feature space rather than coordinate space, and Pointnet does not perform hierarchical pooling, thus Pointnet++ is a closer point cloud generalization of the 2D CNN approach used in \cite{Gatys2016}. In mesh and point cloud representations, there is no information regarding local grouping of samples, thus it is not possible to apply face-wise re-centering, so we use instance normalization for the extracted activations throughout.

For comparison against SOTA, knowing of no existing unsupervised B-Rep style learning methods, as a baseline we use the geometric style embedding of PSNet \cite{Cao2020}, without the colour inputs, which we refer to as PSNet*. PSNet performs geometric and colour style transfer on point clouds without surface normals. 
Its architecture allows us to pre-train its encoder using point cloud reconstruction in a completely unsupervised settings rather than content classification as proposed.
See \autoref{app:models} for further details.
\subsection{Measuring Style Signal}

\begin{figure}[htb]
\begin{center}
   \includegraphics[width=\linewidth]{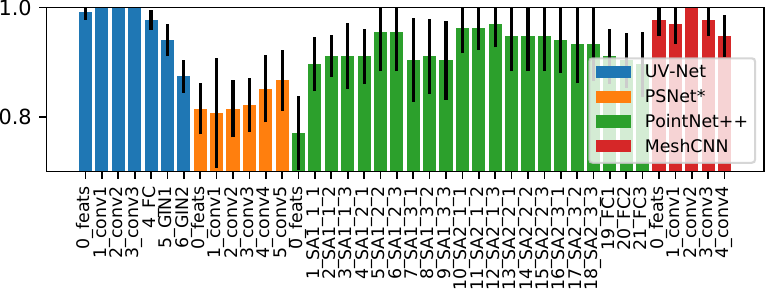}
\end{center}
   \caption{Linear probe classification accuracy scores for each encoder using font labels for evaluation (no font labels used during pre-training). All fonts used here are previously unseen by the networks. Random baseline: 0.25.}
\label{fig:layer_probe_comparison}
\end{figure}

\begin{figure}[htb]
\begin{center}
   \includegraphics[width=.8\linewidth]{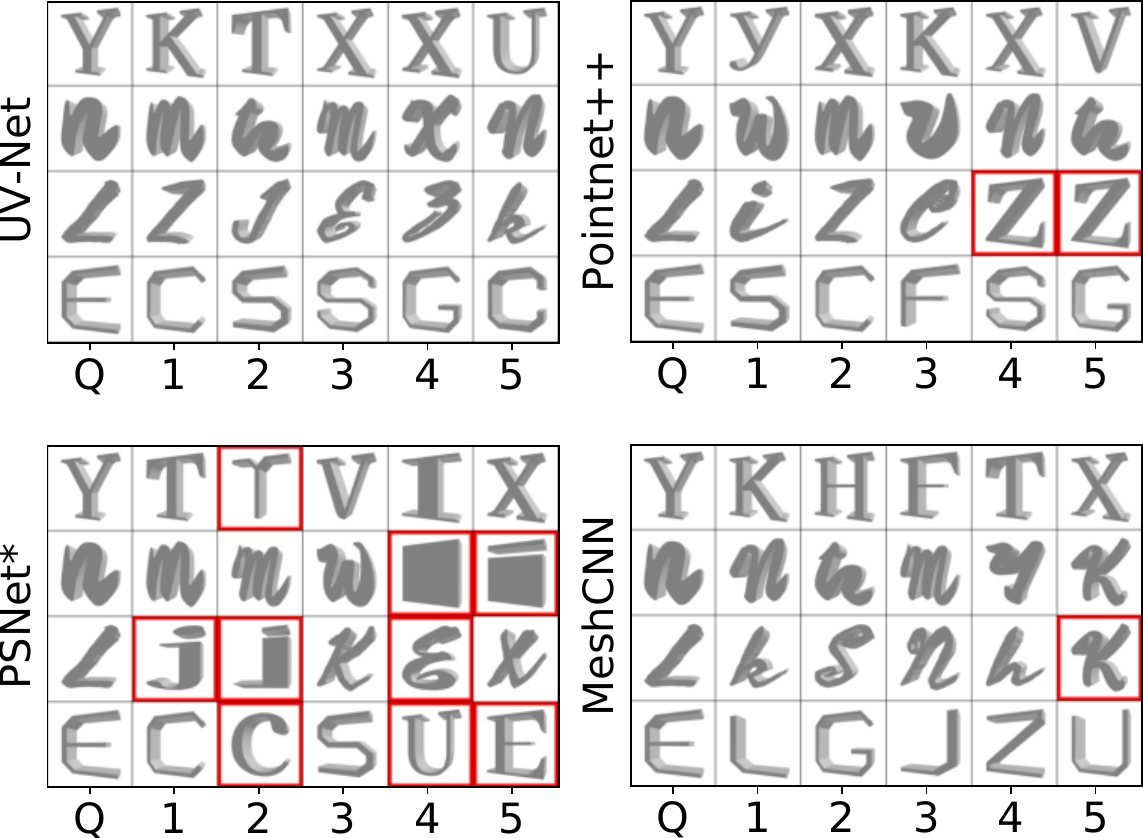}
\end{center}
   \caption{SolidLetters Font Subset: Top-5 queries for a letter from each font, with all weight distributed uniformly over the first $\frac{L}{2}$ layers. Red box indicates result does not match query font.}
\label{fig:topk_equal_weights}
\end{figure}

\begin{figure}[hbt]
\begin{center}
   \includegraphics[width=.8\linewidth]{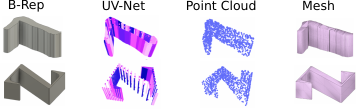}
\end{center}
   \caption{Visualization illustrating the sampling bias advantage of UV-Net, whereby the details in the long surfaces of the `L' are sampled more densely (each face in the B-Rep is sampled with a uniform 10x10 grid) than the simple flat surface of the `Z' making it much easier to differentiate between the different styles than with the uniformly sampled point cloud.}
\label{fig:sample_bias}
\end{figure}

\begin{figure}[htb]
\begin{center}
   \includegraphics[width=\linewidth]{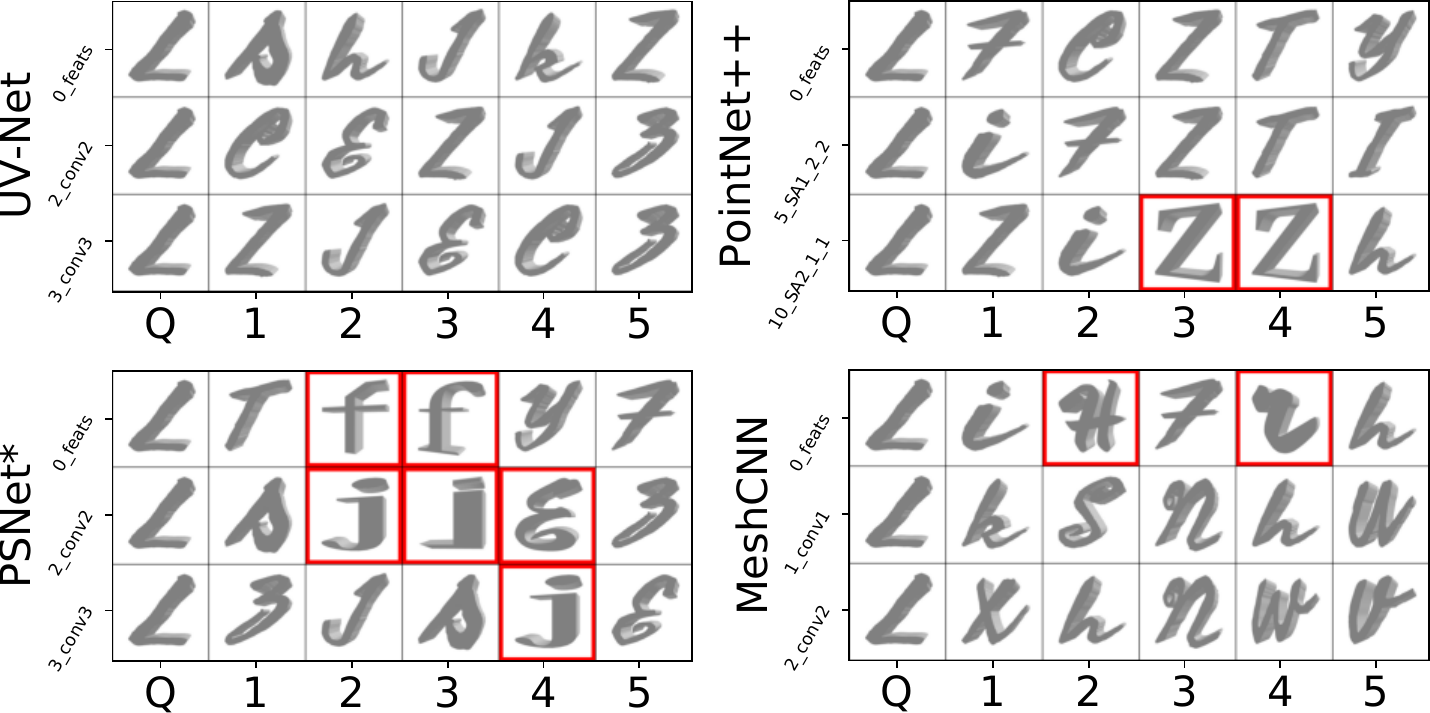}
\end{center}
   \caption{SolidLetters Font Subset: Top-5 queries for the same letter for $l=0$, $l\approx\frac{L}{4}$, and $l\approx\frac{L}{2}$. Red box indicates result does not match query font.}
\label{fig:low_mid_high}
\end{figure}

We adopt the Linear Probe methodology \cite{Alain2016} to measure the amount of style signal present in the Gram matrices of each layer of the pre-trained network. We train a linear classifier on each layer's Gram matrix $G_l$ with ground truth font labels on a subset of the SolidLetters test set. We select four visually distinct fonts in order to strengthen the style labels with respect to style over style compatibility (see \autoref{fig:viaoda_libre_egs}), and due to many fonts in the test set containing almost identical variants. Each encoder is pre-trained with only letter classes as labels, and the four test fonts used in this evaluation are previously unseen. Since the dimensions of the Gram matrices are very large (\ie in some cases $> 2^{19}$), but we have only 137 examples, we perform logistic regression with L2 regularization and 5-fold cross-validation to prevent overfitting. We report the mean validation accuracies.

\autoref{fig:layer_probe_comparison} shows the mean validation accuracy using the extracted Gram matrices from each layer in all four pre-trained models. Compared to random baseline at 0.25, we observe significant indication of style being present in the signals extracted from all layers (including features) for all models. For UV-Net we see the greatest amount of style information in the lowest layers, with the signal reducing deeper into the network. This aligns with our assumption that second order activation statistics transition from style to content representations as network depth increases, as shown for 2D images in \cite{Gatys2016,Karras2019}.

For a qualitative evaluation of our design choices, we perform a top-k query for an example from each font distributing all weight uniformly over the first $\frac{L}{2}$ layers. As shown in \autoref{fig:topk_equal_weights}, with this particular style definition, the style features provided by the pre-trained Pointnet++ model suggest the `Z' from another font is close in style to the query `L', while all UV-Net query results match the target font, and MeshCNN makes only one less obvious mistake. PSNet* has the highest number of errors. 

We hypothesize that this result may be partly due to the sampling strategy of each method. As \autoref{fig:sample_bias} illustrates, UV-Net samples a fixed size grid for each face, thus large faces (such as the long diagonal stem of the `Z') will contribute less to the style features extracted than in PSNet* and Pointnet++ where the point cloud is sampled with uniform density. Therefore, the large diagonal faces have larger influence with Pointnet++ features as network depth increases. Lack of a CNN hierarchy and surface normal inputs may explain lower performance of PSNet* vs PointNet++.

\autoref{fig:low_mid_high} shows the top-k query for the same letter `L' using the style distance from single layers ($l=0$, $l\approx\frac{L}{4}$, and $l\approx\frac{L}{2}$). Supporting our hypothesis above, we see that in this particular scenario, the font is better matched by Pointnet++ in the lower layers. Within the first layer of the network, the features extracted will contain more information about low level structure, \ie bumpy rather than smooth surfaces. Interestingly, for $l \le \frac{L}{2}$, MeshCNN performs worst with the features ($l=0$). We hypothesise this is due to the rotation and scale invariance in the MeshCNN features, whereas UV-Net/Pointnet++ features contain global information.

Finally, comparing with the computational costs of PSNet*, PointNet++, and MeshCNN we observe that the UV-Net encoder with only 645K parameters is 23, 85, and 96 percent faster for style inference, and the Gram matrices require  82, 94, and 35 percent less memory per solid respectively. Full details in \autoref{app:models}. Based on the above results and computational costs, we perform further experiments using the UV-Net encoder only.

\subsection{Gradient Visualization}

\begin{figure}[hbt]
     \centering
        \includegraphics[width=\linewidth]{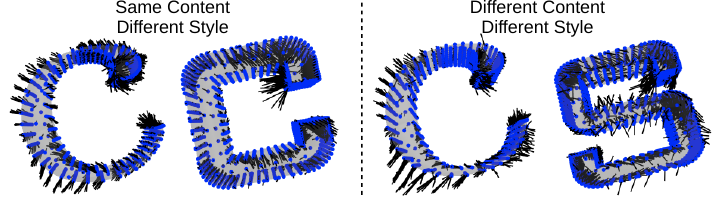}
        \caption{Gradient visualizations of $\mathcal{D}_{style}$ loss (Eq. (\ref{eq:style_distance})) with uniform weight on the first 4 layers (including features), \ie $\vec{w} = [\frac{1}{4}, \frac{1}{4}, \frac{1}{4}, \frac{1}{4}, 0, 0, 0]^{\top}$. Black lines show $-k \cdot \triangledown_{xyz}$, \ie the direction in which to move the point to match the style between the pair.}
        \label{fig:gradient_vis}
\end{figure}

In \autoref{fig:gradient_vis} we visualise our proposed pairwise style distance metric for each B-Rep $\vec{x}$ by computing
\begin{equation}
    \triangledown_{xyz} = \frac{\partial \mathcal{L}_{style}}{\partial \vec{x}_{xyz}}
        \in \R^{N_0 \times 3}
\end{equation}
where $N_0$ is the grid size (number of unmasked UV samples), and $\vec{x}_{xyz}$ is the absolute positions of the UV samples.
For easy interpretation, we plot the vectors $- k \cdot \triangledown_{xyz}$ centered at the samples $\vec{x}_{xyz}$ with black lines to indicate the direction in which a UV sample point should be displaced in order to better match the style between the pair, and $k$ is a constant scaling factor that aids visualization.

In \autoref{fig:gradient_vis} (left) we fix the content and compare different styles. The xyz gradients suggest that the samples of the left example should be moved outwards to match the squarish style on the right, and the samples of the example on the right should be moved inside the solid to match the curves on the left. \autoref{fig:gradient_vis} (right) confirms our approach is able to disentangle style from content, as we compare different content and different style. The gradients on the left example are similar to (left), confirming that the style is matched despite a different content example to compare with.

\subsection{Few-shot Learning of User-Defined Style Measure}

We evaluate few-shot learning of our user-defined style loss on the complete unseen test set, by measuring the mean Precision@10 for each example from a selected font for a range of number of positive and negative user-selected examples. We evaluate on 6 visually distinct fonts (see \autoref{app:few_shot}). Precision@10 is calculated as the proportion of top-10 neighbors that match the target font. For baseline, we compare against the mean Precision@10 with uniform layer weights (one positive and no negative examples). For computational reasons, we reduce the dimensions of each layer's representation $G_l$ to $\min(d_l, 70)$ using PCA.

As shown in \autoref{fig:viaoda_libre_egs}, the font name provides only a weak style label, and as such we are concerned more with the improvement in the mean Precision@10 score than the absolute values. We also consider upper and lower case within the same font as separate labels to further strengthen the associated label, yet also increasing the difficulty of the task as the number of classes doubles to 756.

Positive examples are randomly drawn from the same font and case, and negatives are drawn randomly from all remaining examples. For each number of positives and negatives we perform 20 trials (different positives and negatives each time). We report the mean Precision@10 across all positive examples across all trials, \ie for each number of positives and negatives, every example of a chosen font is queried and evaluated, and this process is repeated 20 times.

\begin{figure}[hbt]
     \centering
     \begin{subfigure}[h]{.35\linewidth}
         \centering
         \includegraphics[width=\linewidth]{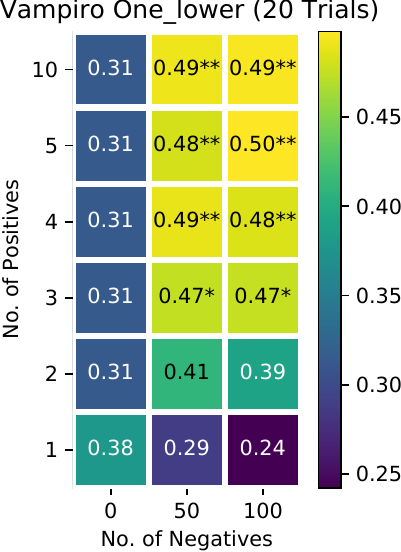}
     \end{subfigure}%
     \qquad
     \begin{subfigure}[h]{.35\linewidth}
         \centering
         \includegraphics[width=\linewidth]{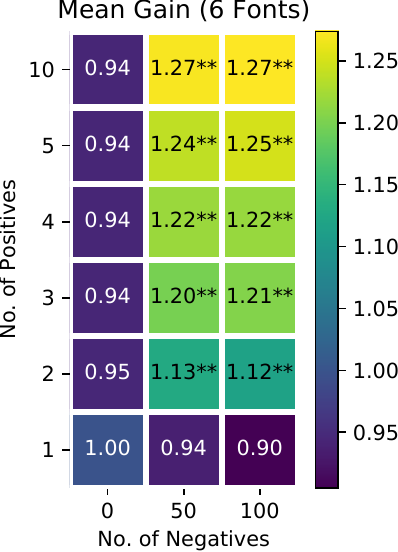}
     \end{subfigure}
        \caption{Left: Mean Precision@10 score for each example of the font after few-shot learning of $\vec{w}^\star$ given a range of number of positive and negative examples. Right: Mean gain in Precision@10 (ratio to baseline) for a selection of fonts. 1 positive + 0 negatives provides baseline using uniform weights. */** indicate a 10\%/5\% statistically significant improvement over baseline respectively.}
        \label{fig:solidmnist_user_optimization}
\end{figure}

\autoref{fig:solidmnist_user_optimization} (left) shows the result for a single font, and (right) the mean gain in Precision@10 (ratio to baseline) for a selection of fonts (further results in \autoref{app:few_shot}). For all combinations of number of positives and negatives greater than 0, we observe a significant improvement in the mean Precision@10 score over the uniformly weighted baseline. Moreover, since negatives are selected randomly from the remaining dataset, we also confirm that providing only positive examples is sufficient to obtain a significantly improved style measure based on the end-user's requirements.

\subsection{Unsupervised Pre-training}

The advantage of our method over existing approaches is that it may be used in unsupervised settings. This is particularly important for B-Reps, since there are no publicly available B-Rep datasets with style labels. We evaluate our approach using the ABC dataset, which contains no content or style labels. For the UVStyle-Net/PSNet* encoder pre-training we use an auto-regressive approach with point cloud reconstruction \cite{Jayaraman2020}. Again, we reduce the dimensions of the style representations $G_l$ to $\min(d_l, 70)$ using PCA.

\begin{figure}[hbt]
\begin{center}
   \includegraphics[width=\linewidth]{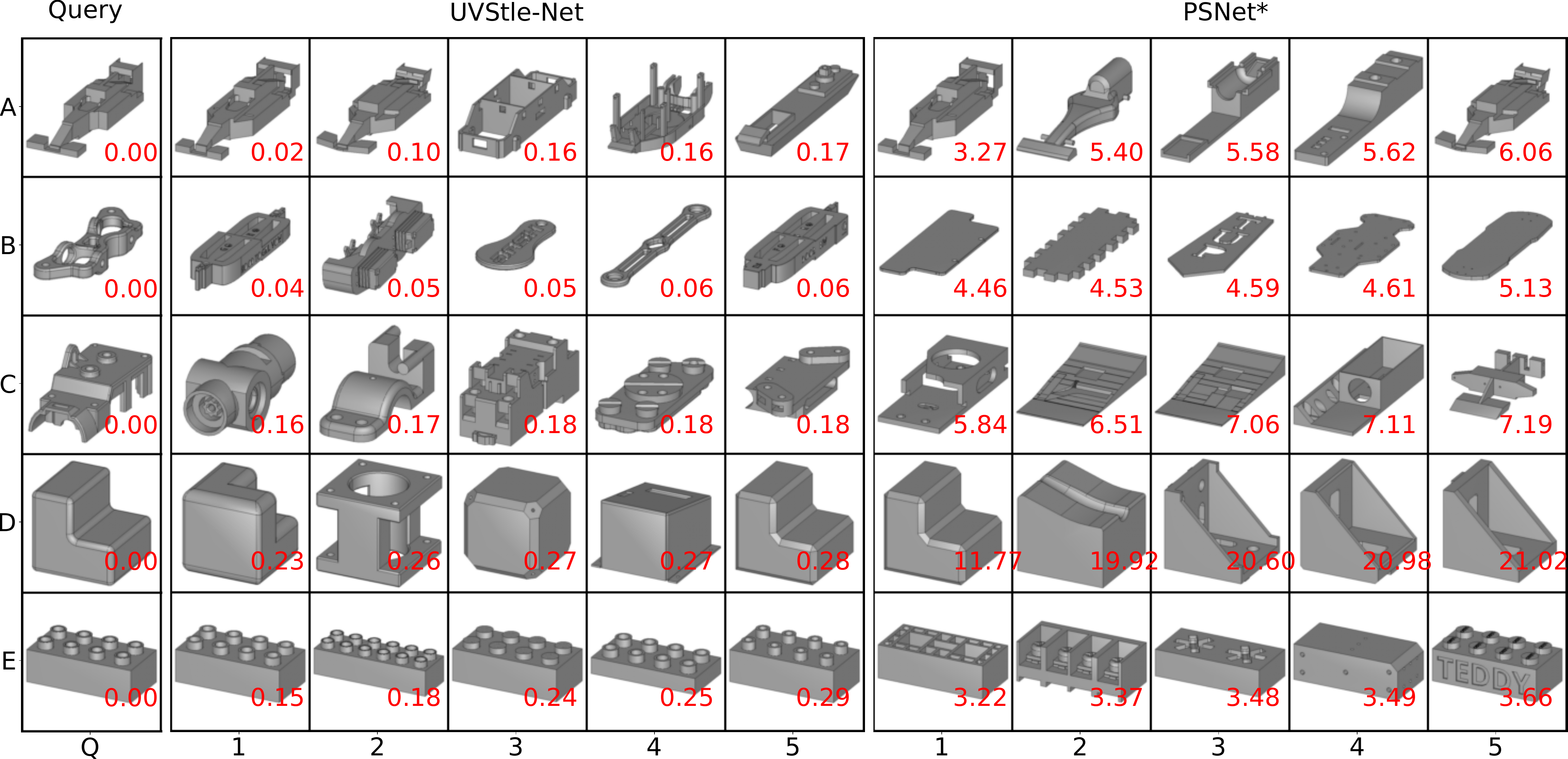}
\end{center}
   \caption{Top-5 query results for ABC dataset from UVStyle-Net and PSNet* pre-trained (unsupervised) with point cloud reconstruction. For UVStyle-Net $\vec{w} = [\frac{1}{3}, \frac{1}{3}, \frac{1}{3}, 0, 0, 0, 0]^\top$. (We recommend zooming in to see stylistic details such as bevels/fillets.)}
\label{fig:abc_top_k}
\end{figure}

\autoref{fig:abc_top_k} shows a few top-5 queries in the style embedding space, using only the lowest 3 layers. For PSNet* queries we use Euclidean distance as this is the metric optimised in \cite{Cao2020}. We observe that UVStyle-Net matches surface style with more variation in content, while in many cases PSNet* matches shapes that roughly occupy the same regions in space as the query, \ie the content. For example, in A: UVStyle-Net finds solids with matching flat surfaces/angles, while PSNet* finds many curved surfaces not present in the query, in C: UVStyle-Net finds more solids with matching curved surfaces, and in E: UVStyle-Net finds blocks with the matching notch style (even with different block size or numbers of notches), whereas PSNet* matches similar sized blocks without the notched style. Comparison of same queries with different UVStyle-Net weights, and PSNet* distance measures are provided in \autoref{app:abc_queries}.

Evaluating our few-shot user-defined style measure, \autoref{fig:overview} shows the nearest neighbour queries for a given target after optimizing the style loss for the user selected examples shown. Selecting filleted solids for positive and a bevelled solid for negative improves the nearest neighbours to the target by pushing away the nearest neighbour of (a) which matches closely in content but not the filleted style. We provide further results in \autoref{app:abc_queries}.

For quantitative evaluation on a real-world dataset, we use subsets of ABC for which we manually curate style labels (details in \autoref{app:abc_labels}). For each model we perform logistic regression on the extracted style embeddings from the pre-trained encoders (Grams from all layers concatenated together). Again we train the encoders using point cloud reconstruction on the complete ABC dataset. We perform 5-fold cross validation with L2 regularization and report the mean validation weighted F1 scores for the best parameters, summarised in \autoref{tab:abc_quant} showing UVStyle-Net significantly outperforms PSNet* on all subsets.

\begin{table}[htb]
    \centering
    \begin{tabularx}{\linewidth}{Xrrr}
    \toprule
    ABC Subset & UVStyle-Net & PSNet* \\
    \midrule
      Flat/Electric &  $\mathbf{0.789 \pm 0.034^*{}^*}$ &  $0.746 \pm 0.038$ \\
 FreeForm/Tubular &  $\mathbf{0.839 \pm 0.011^*{}^*}$ &  $0.808 \pm 0.023$ \\
Angular/Rounded &  $\mathbf{0.805 \pm 0.010^*{}^*}$ &  $0.777 \pm 0.020$ \\
    \bottomrule
    \end{tabularx}
    \caption{Weighted F1 scores for each manually labeled styles subset of ABC. $^*{}^*$ indicates 5\% statistical significance.}
    \label{tab:abc_quant}
\end{table}

\subsection{Ablation}

\begin{figure}[hbt]
\begin{center}
   \includegraphics[width=.8\linewidth]{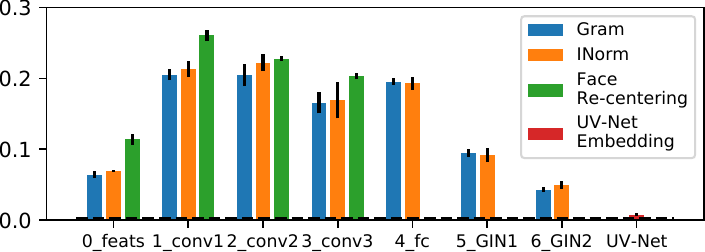}
\end{center}
   \caption{Linear probe scores on complete SolidLetters test set with and without instance/face normalization. Dashed line indicates random classifier baseline.}
\label{fig:ablation_inorm}
\end{figure}

\autoref{fig:ablation_inorm} illustrates the impact of face re-centering and instance normalization using the complete SolidLetters unseen test set. Adopting the linear probe methodology as above, we compare the mean classification accuracy of each layer for predicting all fonts using 5-fold cross-validation. While instance normalization is tested on all layers, face re-centering is not possible beyond the third convolution layer since each face is already represented by a single vector.

The significantly higher scores in the lower layers (excluding features) confirms our assumption that style transitions into content deeper in the network. We also see empirical justification for the use of instance normalization, and in particular face re-centering, which is not possible when working with meshes or point clouds. Comparison with the UV-Net content embedding shows that any of the layer-wise style representations ($G_l$) as proposed in our method are better suited to capturing style information.

\begin{figure}[htb]
\begin{center}
   \includegraphics[width=.8\linewidth]{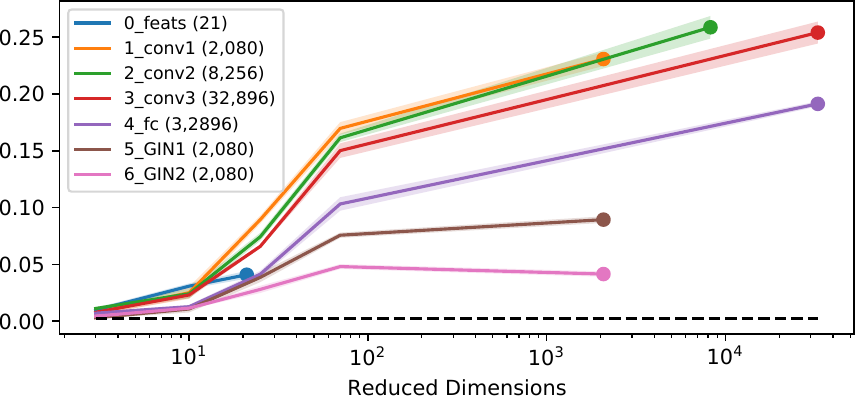}
\end{center}
   \caption{Linear probe scores for each UV-Net layer on complete SolidLetters test set as number of dimensions are reduced. Original dimensions shown in parentheses and marked with $\bullet$.}
\label{fig:dimension_reduction}
\end{figure}

\autoref{fig:dimension_reduction} shows the effect of PCA on the layer-wise style representations ($G_l$) to test the significance of style as a source of variation in each layer. Again, we use linear probes to quantify the style information. In line with our assumptions, the lowest layers ($l=0 \dots 3$) show the greatest amount of style information when the dimensions are sufficiently low, thus indicating that the font style signals are the most significant source of variance in these layers.

\section{Conclusions and Further Work}

UVStyle-Net is a 3D style similarity measure for B-Reps which caters to an end-user's subjective definition of style through few-shot learning based on user selected examples and an unsupervised pre-trained encoder. As a data-driven style measure for B-Reps, which does not require style or content labels yet is adaptable to end-users' requirements, our approach is unique from all existing methods.

Using the SolidLetters font labels for evaluation, our results have demonstrated the applicability of 2D image style principles and assumptions for 3D shapes, and quantified the advantages of our method with B-Reps over alternative methods on meshes and point clouds. In particular, we have confirmed that second order statistics of 3D encoder activations in the first few layers contain style information as the greatest source of variance. We have also shown that our method generates meaningful style gradients, and that the UV-Net sampling strategy and leveraging the face boundary information unique to B-Reps, particularly through face re-centering, significantly improves the style measure.

For a range of 3D fonts and real-world CAD models, we have demonstrated that our proposed method for few-shot learning of user-defined style is effective in improving the style measure for a specific task, even with a minimal number of positive (and optionally, negative) examples. We also demonstrate the benefits of our approach over an existing SOTA method on the real-world ABC dataset where even content labels are not available for encoder pre-training.

A limitation of our method can be seen when solids have very similar content, thus may be improved by stronger disentanglement of style from content. We hypothesize that other unsupervised methods for the encoder pre-training may capture greater detail in the network activations, and therefore improve the style measure on very similar content. We also observe that the current formulation of the few-shot learning often puts all weight on one layer. For future work, we propose investigation into regularization of the few-shot user loss and further investigation into sophisticated distance measures for comparing feature distributions, as well as the natural next step of B-Rep style transfer.

%% file: appendix.tex
\clearpage
\appendix

\section{Introduction to B-Reps}
\label{app:breps}

\begin{figure}[htb]
    \centering
    \includegraphics[width=.7\linewidth]{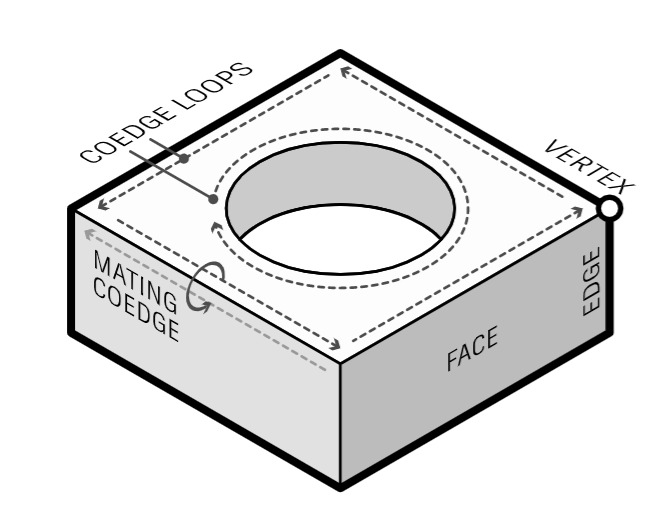}
    \caption{The B-Rep data structure: Faces are defined by parametric surfaces, bounded by loops of trimming curves. Each trimming curve is owned by a topological entity called a coedge, which stores adjacency relationships between faces. Figure from \cite{Weiler1986}.}
    \label{fig:brep}
\end{figure}

B-Reps are loosely analogous to 2D Scalable Vector Graphics (SVGs) for 3D. The precise implementation details vary between different CAD softwares, below we describe the general principles relevant to all B-Reps.

As shown in \autoref{fig:brep}, B-Reps are collections of parametric curves and surfaces along with topological information which describes the adjacency relationships between them \cite{Weiler1986}.
They are typically used to describe closed volumes (solids), but can also represent 2D manifolds (sheets) and curve networks (wire bodies).
Each face of a B-rep body is defined by a parametric surface which is divided into ``visible" and ``hidden" regions by a series of trimming loops. The loops comprise an ordered cycle of coedges, which store pointers to ``mating" coedges on adjacent faces. The loop ordering and coedge-coedge adjacency information provides a full description of the body's topology, while the parametric curves and surfaces provide the geometric information \cite{Masuda}.

B-Reps differ from point clouds and meshes since they are precise representations with continuous smooth surfaces and edge curves --- they are not sampled/discrete. Consequently, complex solids may be expressed with low memory requirements without loss of detail \cite{Lee2001}.

For further information see \cite{Weiler1986,Masuda,Lee2001}.

\section{Few Shot Learning}
\label{app:few_shot}

\begin{figure}[!h]
    \includegraphics[width=.8\linewidth]{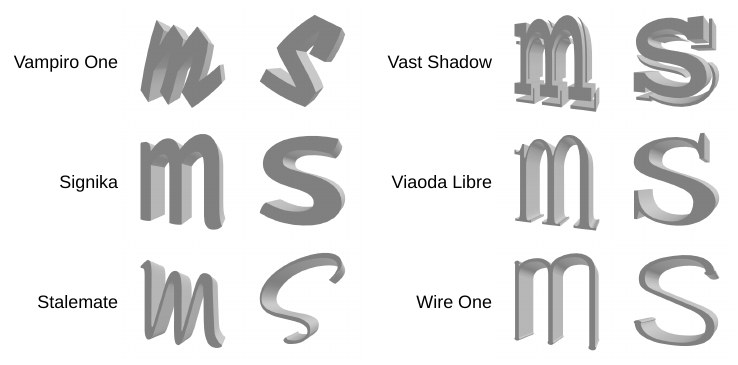}
  \caption{}
\label{fig:optimization_egs}
\end{figure}

\begin{figure}[!h]
\centering
\begin{subfigure}[h]{.35\linewidth}
   \includegraphics[width=\linewidth]{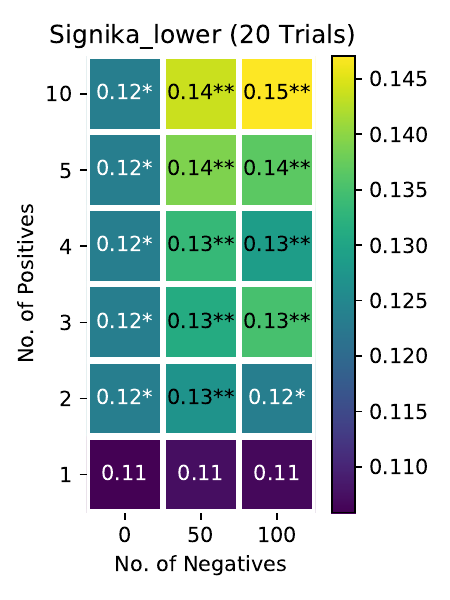}
\end{subfigure}%
\begin{subfigure}[h]{.35\linewidth}
   \includegraphics[width=\linewidth]{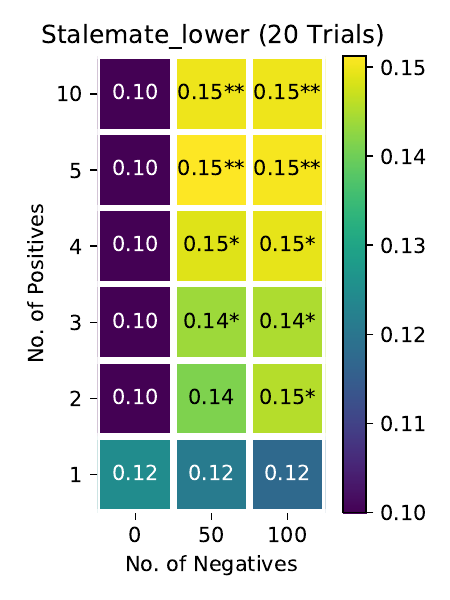}
\end{subfigure}
\begin{subfigure}[h]{.35\linewidth}
   \includegraphics[width=\linewidth]{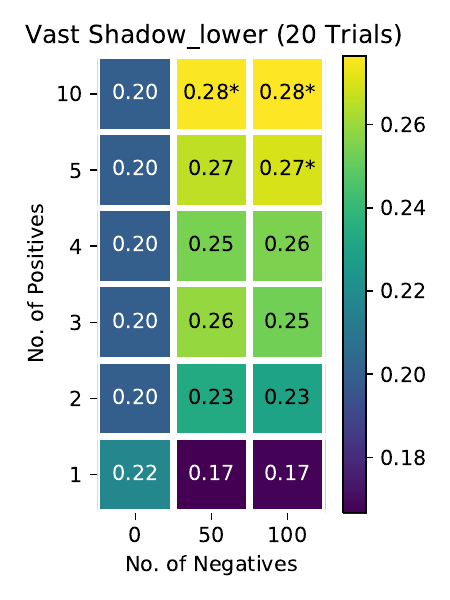}
\end{subfigure}%
\begin{subfigure}[h]{.35\linewidth}
   \includegraphics[width=\linewidth]{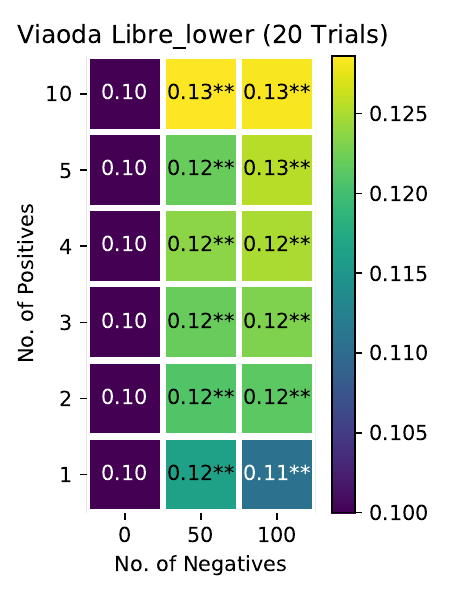}
\end{subfigure}
\begin{subfigure}[h]{.35\linewidth}
   \includegraphics[width=\linewidth]{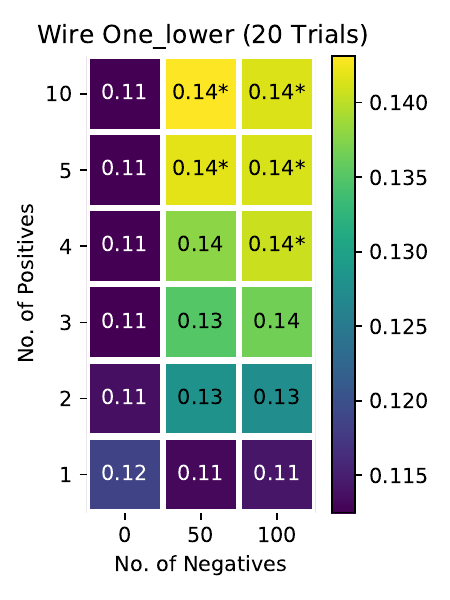}
\end{subfigure}
  \caption{}
\label{fig:fewshot_a}
\end{figure}

\autoref{fig:fewshot_a} shows the absolute mean Precision@10 scores over a range of number of positive and negative examples of each of the unseen fonts we tested. These in conjunction with the font shown in \autoref{fig:solidmnist_user_optimization} (left) are used to calculate the mean gain shown in \autoref{fig:solidmnist_user_optimization} (right). Examples of each font are given in \autoref{fig:optimization_egs}. 1 positive and 0 negative indicates baseline using equal layer weights.

For the most visually distinct fonts (\ie `Vampiro One' and `Vast Shadow'), the equal weights baseline is highest. The amount of improvement is dependant on the self-consistency of style within the font and the number of similar fonts in the test set. We observe greater self-consistency within `Vampiro One' and `Vast Shadow' while being distinct from the rest of the test set. While the other fonts still show improvement, we expect lower results due to their inconsistency or lack of distinct stylistic features, \ie in `Stalemate' the `m' and `s' appear to be stylistically compatible, but the max curvatures of the `m' are much greater than in the `s' - the style is not obviously the same.

\clearpage
\section{Unsupervised Pre-training}
\label{app:abc_queries}

\begin{figure}[!h]
\begin{center}
   \includegraphics[width=\linewidth]{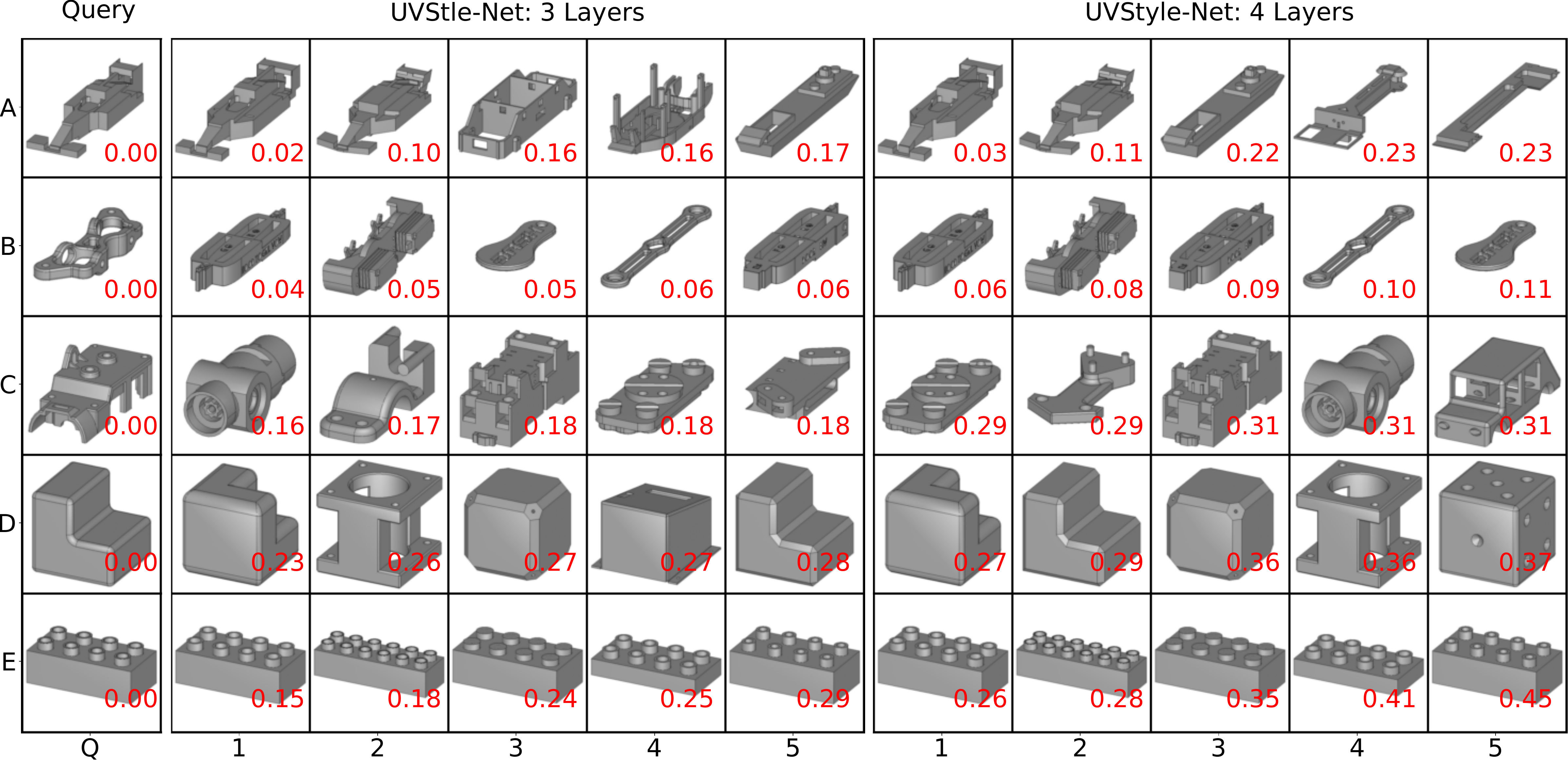}
\end{center}
   \caption{Comparison of top-5 queries with different weights for UVStyle-Net on ABC dataset with unsupervised pre-training. 3 Layers: $\vec{w} = [\frac{1}{3}, \frac{1}{3}, \frac{1}{3}, 0, 0, 0, 0]^\top$, 4 Layers: $\vec{w} = [\frac{1}{4}, \frac{1}{4}, \frac{1}{4}, \frac{1}{4}, 0, 0, 0]^\top$.}
\label{fig:abc_extra_lower}
\end{figure}

\begin{figure}[!h]
\begin{center}
   \includegraphics[width=.6\linewidth]{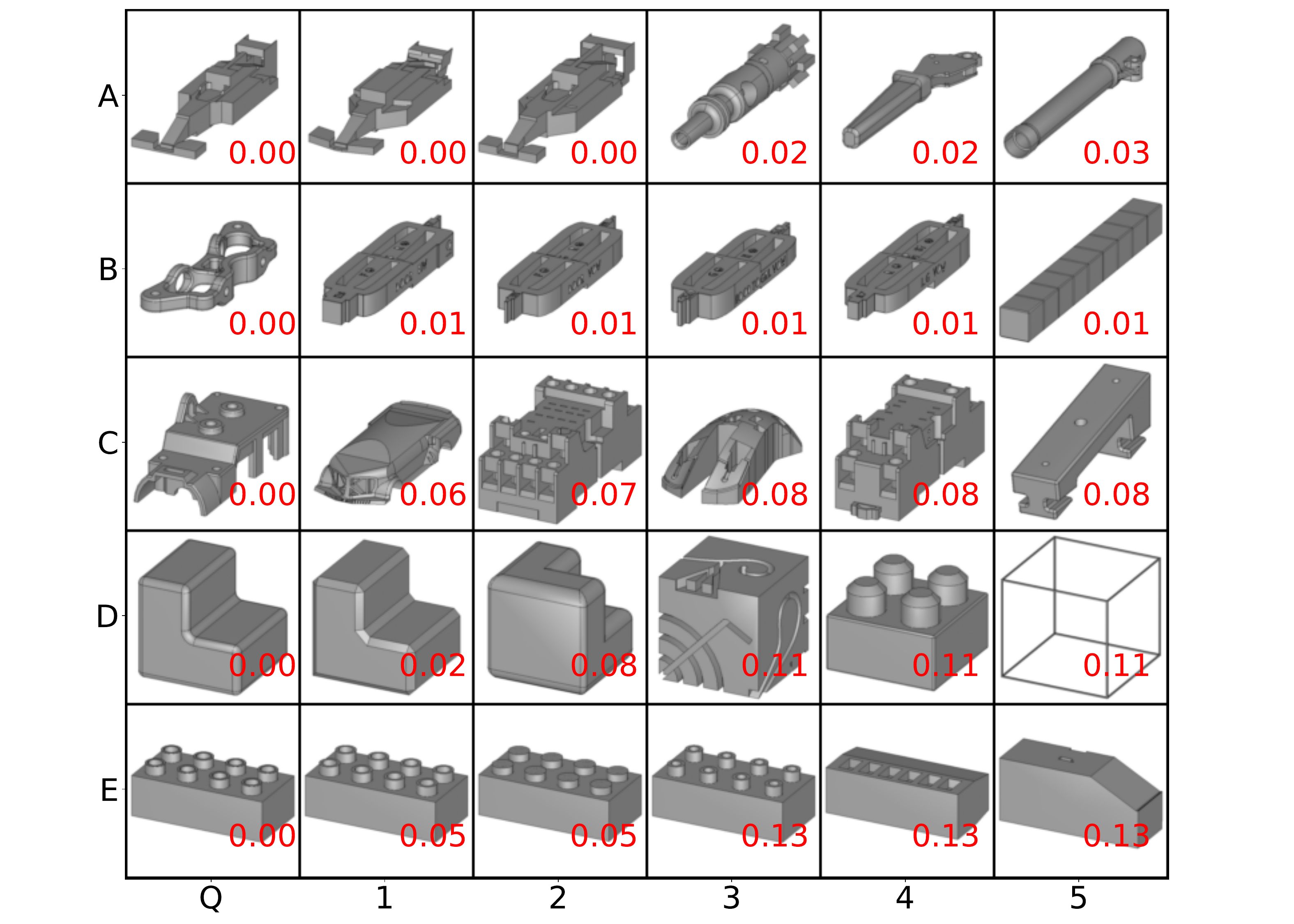}
\end{center}
   \caption{Top-5 query results for ABC dataset from UVStyle-Net with unsupervised pre-training. $\vec{w} = [0, 0, 0, 0, 0, 0, 1]^\top$. Weighting the upper layers of the network moves the definition of style closer to content, where the distance measure is more about the general shape and size and global features, and less about the fine details and local features.}
\label{fig:abc_top_layers}
\end{figure}

\begin{figure}[!h]
\begin{center}
   \includegraphics[width=\linewidth]{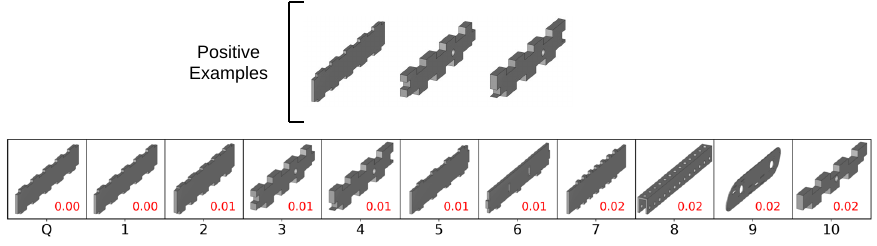}
\end{center}
   \caption{Optimizing $\mathcal{L}_{user}$ with positive examples matching in content results in layer weight distributed over the upper layers. $\vec{w}^\star \approx [0, 0, 0, 0, 0, \frac{1}{3}, \frac{2}{3}]^\top$.}
\label{fig:abc_content_optimization}
\end{figure}

\begin{figure}[ht]
    \centering
    \includegraphics[width=.6\linewidth]{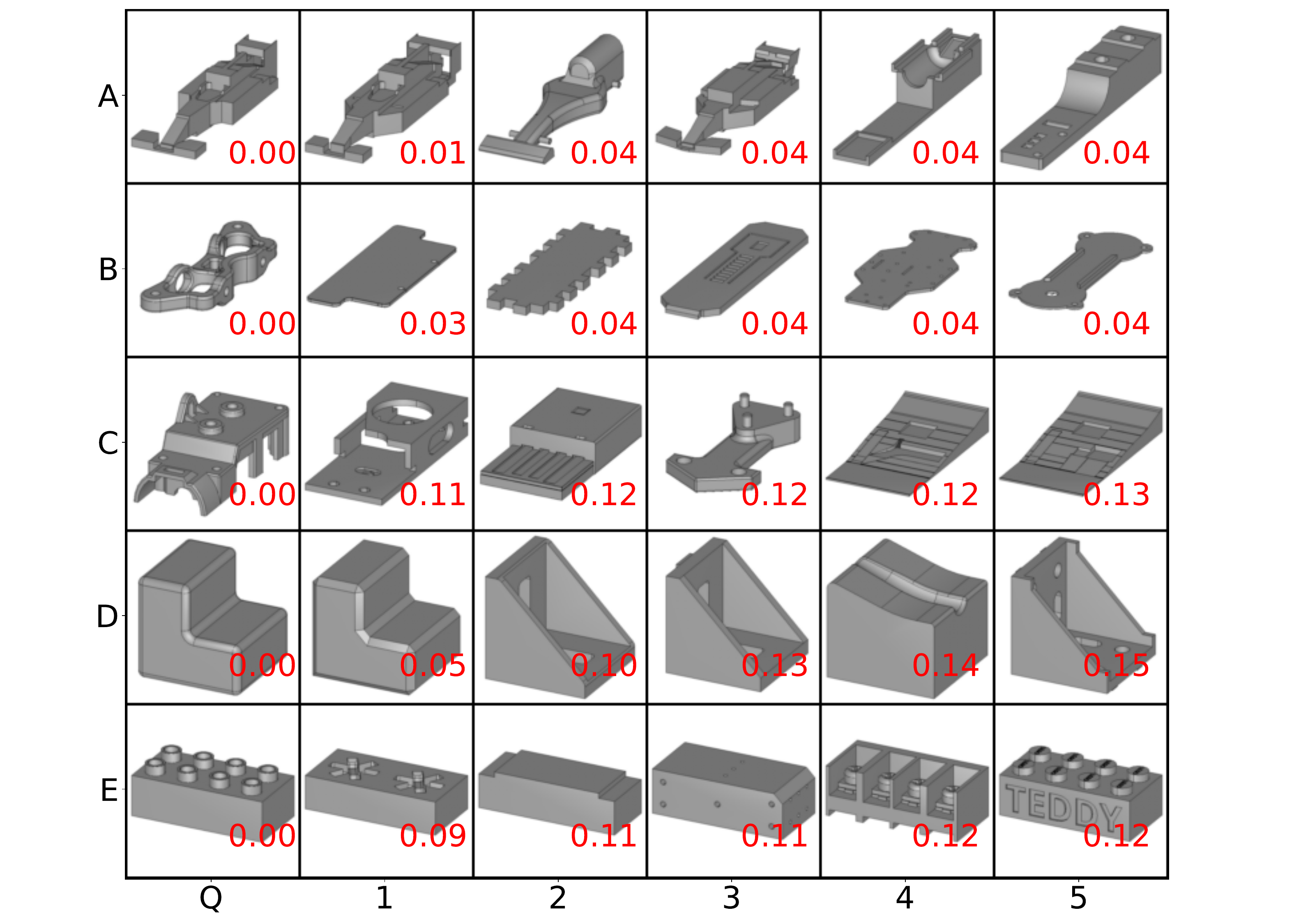}
    \caption{Top-5 queries for PSNet* with cosine distance.}
    \label{fig:my_label}
\end{figure}

\section{ABC Style Labels}
\label{app:abc_labels}

\begin{figure}
    \centering
    \includegraphics[width=.6\linewidth]{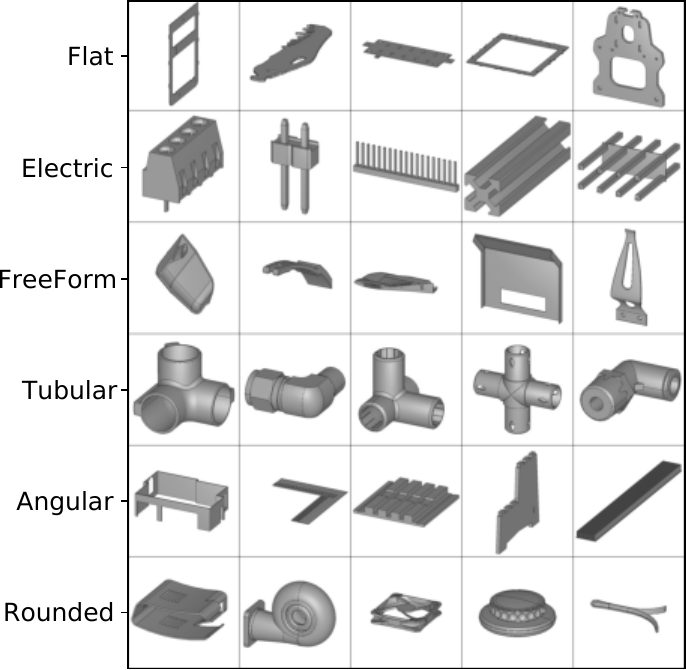}
    \caption{Examples of each ABC style subset classes. Each style is selected to be visually distinct, and while some classes contain the same types of objects, \ie, `Tubular', the overall shapes (the content) are diverse.}
    \label{fig:abc_quant_egs}
\end{figure}

There is a fundamental lack of publicly available labeled B-Rep data, with no existing B-Rep datasets containing style labels. To enable quantitative evaluation of our method and promote further work in this area we contribute a set of manually assigned style labels for a subset of the ABC solid models. We selected categories with distinct styles while containing diverse content. Examples of each category are shown in \autoref{fig:abc_quant_egs} and details of the class sizes in \autoref{tab:abc_classes}. These labels are available at \href{https://github.com/AutodeskAILab/UVStyle-Net}{github.com/AutodeskAILab/UVStyle-Net}.

\begin{table}[tbp]
    \centering
    \begin{tabular}{ll}
    \toprule
    ABC Subset & Examples \\
    \midrule
    Flat/Electric   & 389/58 \\
    Free Form/Pipe  & 241/24 \\
    Angular/Rounded & 834/106 \\
    \bottomrule
    \end{tabular}
    \caption{Manually labeled ABC style subsets.}
    \label{tab:abc_classes}
\end{table}

\section{SolidLetters Test Set Generation}
\label{app:generation}

\begin{figure}
    \centering
    \begin{subfigure}[b]{.45\linewidth}
       \includegraphics[width=\linewidth]{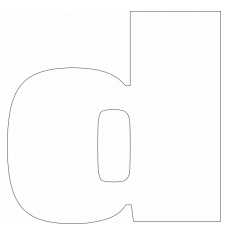}
       \subcaption{2D Font wire}
    \end{subfigure}%
    \begin{subfigure}[b]{.45\linewidth}
       \includegraphics[width=\linewidth]{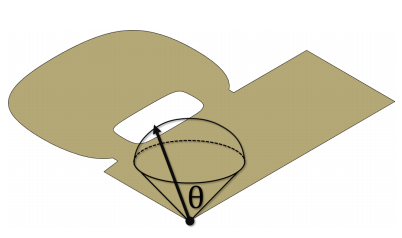}
       \subcaption{Select random extrude angle}
    \end{subfigure}
    \begin{subfigure}[h]{.45\linewidth}
       \includegraphics[width=\linewidth]{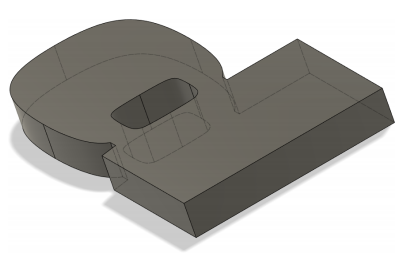}
       \subcaption{Extrude}
    \end{subfigure}%
    \begin{subfigure}[h]{.45\linewidth}
       \includegraphics[width=\linewidth]{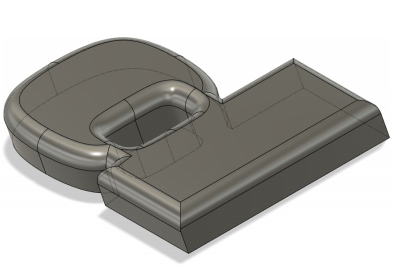}
       \subcaption{Fillet}
    \end{subfigure}%
    \caption{Steps for generation of SolidLetters dataset. For test set, extrude angle and fillet amount are fixed. Figure from \cite{Jayaraman2020}}
    \label{fig:solid_gen}
\end{figure}

For SolidLetters, the training data is generated as per \cite{Jayaraman2020} using code and font wires provided by the authors. The key steps are illustrated in \autoref{fig:solid_gen}.

The held-out test set is regenerated to strengthen the associated style labels by removing inconsistent sources of randomness within font classes. The extrusion depth and angle are fixed across all fonts. Filleting size is also fixed, and is applied only to fonts where it possible to apply it to all examples of that font. Filleting is not possible for some examples due to the complexity of the solids. If filleting is unsuccessful on any example, all examples of that font are left without fillets.

All SolidLetters data used is freely available at \href{https://github.com/AutodeskAILab/UVStyle-Net}{github.com/AutodeskAILab/UVStyle-Net}.

\section{Model Details}
\label{app:models}

For MeshCNN we use the author's code from \url{https://github.com/ranahanocka/MeshCNN}, for Pointnet++ we use \url{https://github.com/erikwijmans/Pointnet2\_PyTorch}. All experiments performed on \href{https://aws.amazon.com/about-aws/whats-new/2017/10/introducing-amazon-ec2-p3-instances/}{AWS p3.2xlarge}.

\autoref{tab:metainfo} shows details about the model hyper parameters and meta information. For MeshCNN, we remeshed the data to 15000 edges per solid and for Pointnet++ we used their multi-scale grouping (MSG) setup. Other parameters and architecture choices not mentioned here, are set to default. All point clouds are sampled with 1024 points.

For PSNet* we use the Pointnet implementation from \url{https://github.com/WangYueFt/dgcnn} and extract the Gram matrices from the first 4 layers as detailed in \cite{Cao2020}. While PSNet works with geometry and colour, we use only the geometric part in our comparisons.

In \autoref{tab:modelcompare} we compare the computational costs of each encoder.

\begin{table}[!ht]
  \begin{center}
    \begin{tabular}{lrrrrr}
        \toprule
        Model & LR & N & F & BS & Opt\\
        \midrule
        UV-Net & 1e-4  & BN & 7 & 128 & Adam\\
        PSNet* & 1e-4  & BN & 3 & 128 & Adam\\
        Pointnet++ & 1e-3 & BN & 6 & 32 & Adam\\
        MeshCNN & 2e-4 & GN & 5 & 4& Adam \\
        \bottomrule
    \end{tabular}
    \caption{Hyper-parameters and meta information about the models for SolidLetters runs. LR denotes learning rate, N type of norm (\ie batch norm or group norm), F input feature dimension, BS batch size and Opt, the type of optimiser used.}
    \label{tab:metainfo}
  \end{center}
\end{table}

\begin{table}[htb]
  \begin{center}
    \begin{tabular}{lrrrr}
        \toprule
        Encoder & $L$ & Parameters & Time & Size\\
        \midrule
        UV-Net & 7 & \textbf{645,596} & 93min/\textbf{88s} & \textbf{199 KB}\\
        PSNet* & 5 & 813,914 & 165min/115s & 1.08MB\\
        Pointnet++ & 22 & 1,746,420 & \textbf{43min}/603s & 3.32 MB\\
        MeshCNN & 5 & 1,322,982 & 29hr/38min & 305 KB\\
        \bottomrule
    \end{tabular}
    \caption{Comparison of 3D encoder methods. $L$ is total number of layers (including features), times given are pre-training/style inference on complete SolidLetters test set. Size is the memory required for a single style embedding (containing one Gram per layer) for a single solid --- note this is not dependent on the size of the input solid. For style inference UV-Net is the most compute and memory efficient. MeshCNN suffers from small batch size due to necessarily large meshes, and Pointnet++ suffers from larger Gram matrices.}
    \label{tab:modelcompare}
  \end{center}
\end{table}

%% file: iccv.bbl
\begin{thebibliography}{10}\itemsep=-1pt

\bibitem{Ahmed2019}
Eman Ahmed, Alexandre Saint, Abdelrahman Shabayek, and Kseniya Cherenkova.
\newblock {A survey on Deep Learning Advances on Different 3D Data
  Representations}.
\newblock {\em arXiv}, 1(1), 2019.

\bibitem{Alain2016}
Guillaume Alain and Yoshua Bengio.
\newblock {Understanding intermediate layers using linear classifier probes}.
\newblock {\em arXiv}, 2016.

\bibitem{Azadi}
Samaneh Azadi, Matthew Fisher, Vladimir Kim, Zhaowen Wang, Eli Shechtman, and
  Trevor Darrell.
\newblock {Multi-content GAN for Few-Shot Font Style Transfer}.
\newblock In {\em Proceedings of the IEEE Computer Society Conference on
  Computer Vision and Pattern Recognition}, pages 7564--7573, 2018.

\bibitem{Babaeizadeh2019}
Mohammad Babaeizadeh and Golnaz Ghiasi.
\newblock {Adjustable Real-time Style Transfer}.
\newblock {\em Deep Generative Models for Highly Structured Data, DGS@ICLR 2019
  Workshop}, 11 2018.

\bibitem{Bronstein2017}
Michael~M Bronstein, Joan Bruna, Yann LeCun, Arthur Szlam, and Pierre
  Vandergheynst.
\newblock {Geometric Deep Learning: Going beyond Euclidean data}.
\newblock {\em IEEE Signal Processing Magazine}, 34(4):18--42, 7 2017.

\bibitem{Cao2020}
Xu Cao, Weimin Wang, Katashi Nagao, and Ryosuke Nakamura.
\newblock {PSNet: A style transfer network for point cloud stylization on
  geometry and color}.
\newblock {\em Proceedings - 2020 IEEE Winter Conference on Applications of
  Computer Vision, WACV 2020}, pages 3326--3334, 2020.

\bibitem{Cao}
Zhangjie Cao, Qixing Huang, and Ramani Karthik.
\newblock {3D Object Classification via Spherical Projections}.
\newblock In {\em 2017 International Conference on 3D Vision (3DV)}, pages
  566--574. IEEE, 10 2017.

\bibitem{Chang2015}
Angel~X Chang, Thomas Funkhouser, Leonidas Guibas, Pat Hanrahan, Qixing Huang,
  Zimo Li, Silvio Savarese, Manolis Savva, Shuran Song, Hao Su, Jianxiong Xiao,
  Li Yi, and Fisher Yu.
\newblock {ShapeNet: An Information-Rich 3D Model Repository}.
\newblock {\em arXiv}, 2015.

\bibitem{Cohen2017}
Gregory Cohen, Saeed Afshar, Jonathan Tapson, and André van Schaik.
\newblock {EMNIST: an extension of MNIST to handwritten letters}.
\newblock {\em arXiv}, 2 2017.

\bibitem{DeHaan2020}
Pim de Haan, Maurice Weiler, Taco Cohen, and Max Welling.
\newblock {Gauge Equivariant Mesh CNNs: Anisotropic convolutions on geometric
  graphs}.
\newblock {\em arXiv}, 3 2020.

\bibitem{Gatys2016}
Leon~A Gatys, Alexander~S Ecker, and Matthias Bethge.
\newblock {Image Style Transfer Using Convolutional Neural Networks}.
\newblock In {\em 2016 IEEE Conference on Computer Vision and Pattern
  Recognition (CVPR)}, volume 2016-Decem, pages 2414--2423. IEEE, 6 2016.

\bibitem{Griffiths2019}
David Griffiths and Jan Boehm.
\newblock {A Review on Deep Learning Techniques for 3D Sensed Data
  Classification}.
\newblock {\em Remote Sensing}, 11(12):1499, 6 2019.

\bibitem{Guo2019}
Yulan Guo, Hanyun Wang, Qingyong Hu, Hao Liu, Li Liu, and Mohammed Bennamoun.
\newblock {Deep Learning for 3D Point Clouds: A Survey}.
\newblock {\em IEEE Transactions on Pattern Analysis and Machine Intelligence},
  pages 1--1, 2020.

\bibitem{Hanocka2019}
Rana Hanocka, Amir Hertz, Noa Fish, Raja Giryes, Shachar Fleishman, and Daniel
  Cohen-Or.
\newblock {MeshCNN: A Network with an Edge}.
\newblock {\em ACM Transactions on Graphics}, 38(4):1--12, 9 2018.

\bibitem{Huang2017}
Xun Huang and Serge Belongie.
\newblock {Arbitrary Style Transfer in Real-Time with Adaptive Instance
  Normalization}.
\newblock In {\em 2017 IEEE International Conference on Computer Vision
  (ICCV)}, volume 2017-Octob, pages 1510--1519. IEEE, 10 2017.

\bibitem{Jayaraman2020}
Pradeep~Kumar Jayaraman, Aditya Sanghi, Joseph Lambourne, Thomas Davies, Hooman
  Shayani, and Nigel Morris.
\newblock {UV-Net: Learning from Curve-Networks and Solids}.
\newblock {\em arXiv}, 6 2020.

\bibitem{Jiang2020}
Liming Jiang, Changxu Zhang, Mingyang Huang, Chunxiao Liu, Jianping Shi, and
  Chen~Change Loy.
\newblock {TSIT: A Simple and Versatile Framework for Image-to-Image
  Translation}.
\newblock {\em arXiv}, 7 2020.

\bibitem{Karras2019}
Tero Karras, Samuli Laine, and Timo Aila.
\newblock {A Style-Based Generator Architecture for Generative Adversarial
  Networks}.
\newblock {\em 2019 IEEE/CVF Conference on Computer Vision and Pattern
  Recognition (CVPR)}, pages 4396--4405, 12 2018.

\bibitem{Lehtinen}
Tero Karras, Samuli Laine, Miika Aittala, Janne Hellsten, Jaakko Lehtinen, and
  Timo Aila.
\newblock {Analyzing and Improving the Image Quality of StyleGAN}.
\newblock {\em arXiv}, 2019.

\bibitem{KochBerlinskoch}
Sebastian Koch, Albert Matveev, Zhongshi Jiang, Francis Williams, Alexey
  Artemov, Evgeny Burnaev, Marc Alexa, Denis Zorin, and Daniele Panozzo.
\newblock {ABC: A big cad model dataset for geometric deep learning}.
\newblock In {\em Proceedings of the IEEE Computer Society Conference on
  Computer Vision and Pattern Recognition}, volume 2019-June, pages 9593--9603,
  2019.

\bibitem{Lee2001}
Sang~Hun Lee and Kunwoo Lee.
\newblock {Partial entity structure: A compact boundary representation for
  non-manifold geometric modeling}.
\newblock {\em Journal of Computing and Information Science in Engineering},
  1(4):356--365, 2001.

\bibitem{Lim2016}
Isaak Lim, Anne Gehre, and Leif Kobbelt.
\newblock {Identifying Style of 3D Shapes using Deep Metric Learning}.
\newblock {\em Computer Graphics Forum}, 35(5):207--215, 8 2016.

\bibitem{DerekLiu2020}
Hsueh Ti~Derek Liu, Vladimir~G Kim, Siddhartha Chaudhuri, Noam Aigerman, and
  Alec Jacobson.
\newblock {Neural subdivision}.
\newblock {\em ACM Transactions on Graphics}, 39(4):16, 2020.

\bibitem{Liu2015}
Tianqiang Liu, Aaron Hertzmann, Wilmot Li, and Thomas Funkhouser.
\newblock {Style compatibility for 3D furniture models}.
\newblock {\em ACM Transactions on Graphics}, 34(4):1--9, 7 2015.

\bibitem{Lun2015}
Zhaoliang Lun, Evangelos Kalogerakis, and Alla Sheffer.
\newblock {Elements of style: Learning perceptual shape style similarity}.
\newblock In {\em ACM Transactions on Graphics}, volume~34, 2015.

\bibitem{Masuda}
Hiroshi Masuda, Kenji Shimada, Masayuki Numao, and Shinji Kawabe.
\newblock {A Mathematical Theory and Applications of Non-Manifold Geometric
  Modelling}.
\newblock In {\em International Symposium on Advanced Geometric Modelling for
  Engineering Applications}, pages 89--103, 1989.

\bibitem{Pan2017}
Tse-Yu Pan, Yi-Zhu Dai, Wan-Lun Tsai, and Min-Chun Hu.
\newblock {Deep model style: Cross-class style compatibility for 3D furniture
  within a scene}.
\newblock In {\em 2017 IEEE International Conference on Big Data (Big Data)},
  volume 2018-Janua, pages 4307--4313. IEEE, 12 2017.

\bibitem{Pan2019}
Xiang Pan, Jie Lu, and Fuchang Liu.
\newblock {3D Patch-Based Sparse Learning for Style Feature Extraction}.
\newblock {\em IEEE Access}, 7:172403--172412, 2019.

\bibitem{Park2020}
Taesung Park, Jun-Yan Zhu, Oliver Wang, Jingwan Lu, Eli Shechtman, Alexei~A
  Efros, and Richard Zhang.
\newblock {Swapping Autoencoder for Deep Image Manipulation}.
\newblock {\em arXiv}, 7 2020.

\bibitem{Polania2020}
Luisa~F. Polania, Mauricio Flores, Yiran Li, and Matthew Nokleby.
\newblock {Learning Furniture Compatibility with Graph Neural Networks}.
\newblock {\em arXiv}, 2020.

\bibitem{Qi}
Charles~R Qi, Hao Su, Kaichun Mo, and Leonidas~J Guibas.
\newblock {PointNet: Deep learning on point sets for 3D classification and
  segmentation}.
\newblock In {\em Proceedings - 30th IEEE Conference on Computer Vision and
  Pattern Recognition, CVPR 2017}, volume 2017-Janua, pages 77--85, 2017.

\bibitem{Qi2017}
Charles~R. Qi, Li Yi, Hao Su, and Leonidas~J. Guibas.
\newblock {PointNet++: Deep Hierarchical Feature Learning on Point Sets in a
  Metric Space}.
\newblock {\em Advances in Neural Information Processing Systems},
  2017-Decem:5100--5109, 6 2017.

\bibitem{Segu}
Mattia Segu, Margarita Grinvald, Roland Siegwart, and Federico Tombari.
\newblock {3DSNet: Unsupervised Shape-to-Shape 3D Style Transfer}.
\newblock {\em arXiv}, 11 2020.

\bibitem{Su}
Hang Su, Subhransu Maji, Evangelos Kalogerakis, and Erik Learned-Miller.
\newblock {Multi-view Convolutional Neural Networks for 3D Shape Recognition}.
\newblock In {\em 2015 IEEE International Conference on Computer Vision
  (ICCV)}, volume 2015 Inter, pages 945--953. IEEE, 12 2015.

\bibitem{Ulyanov2017}
Dmitry Ulyanov, Andrea Vedaldi, and Victor Lempitsky.
\newblock {Improved Texture Networks: Maximizing Quality and Diversity in
  Feed-Forward Stylization and Texture Synthesis}.
\newblock In {\em 2017 IEEE Conference on Computer Vision and Pattern
  Recognition (CVPR)}, volume 2017-Janua, pages 4105--4113. IEEE, 7 2017.

\bibitem{Virtanen2020}
Pauli Virtanen, Ralf Gommers, Travis~E. Oliphant, Matt Haberland, Tyler Reddy,
  David Cournapeau, Evgeni Burovski, Pearu Peterson, Warren Weckesser, Jonathan
  Bright, Stéfan~J. van~der Walt, Matthew Brett, Joshua Wilson, K.~Jarrod
  Millman, Nikolay Mayorov, Andrew R.~J. Nelson, Eric Jones, Robert Kern, Eric
  Larson, C.~J. Carey, İlhan Polat, Yu Feng, Eric~W. Moore, Jake VanderPlas,
  Denis Laxalde, Josef Perktold, Robert Cimrman, Ian Henriksen, E.~A. Quintero,
  Charles~R. Harris, Anne~M. Archibald, Antônio~H. Ribeiro, Fabian Pedregosa,
  and Paul van Mulbregt.
\newblock {SciPy 1.0: fundamental algorithms for scientific computing in
  Python}.
\newblock {\em Nature Methods}, 17(3):261--272, 3 2020.

\bibitem{Wang2019b}
Yue Wang, Yongbin Sun, Ziwei Liu, Sanjay~E Sarma, Michael~M Bronstein, and
  Justin~M Solomon.
\newblock {Dynamic Graph CNN for Learning on Point Clouds}.
\newblock {\em ACM Transactions on Graphics}, 38(5):1--12, 11 2019.

\bibitem{Weiler1986}
Kevin~J Weiler.
\newblock {Topological structures for geometric modeling}, 1986.

\bibitem{Xu}
Keyulu Xu, Weihua Hu, Jure Leskovec, and Stefanie Jegelka.
\newblock {How Powerful Are Graph Neural Networks?}
\newblock In {\em ICLR}, 2019.

\bibitem{Zhanga}
Yulun Zhang, Chen Fang, Yilin Wang, Zhaowen Wang, Zhe Lin, Yun Fu, and Jimei
  Yang.
\newblock {Multimodal Style Transfer via Graph Cuts}.
\newblock In {\em 2019 IEEE/CVF International Conference on Computer Vision
  (ICCV)}, volume 2019-Octob, pages 5942--5950. IEEE, 10 2019.

\bibitem{ZhirongWu2015}
{Zhirong Wu}, Shuran Song, Aditya Khosla, {Fisher Yu}, {Linguang Zhang},
  {Xiaoou Tang}, and Jianxiong Xiao.
\newblock {3D ShapeNets: A deep representation for volumetric shapes}.
\newblock In {\em 2015 IEEE Conference on Computer Vision and Pattern
  Recognition (CVPR)}, volume 07-12-June, pages 1912--1920. IEEE, 6 2015.

\end{thebibliography}
